\pdfoutput=1

\documentclass[journal]{IEEEtran}
\usepackage{hyperref}

\usepackage{cite}

\usepackage{graphicx}

\ifCLASSINFOpdf
\else
\fi
\usepackage{amsmath}

\usepackage{multirow}

\ifCLASSOPTIONcompsoc
  \usepackage[caption=false,font=normalsize,labelfont=sf,textfont=sf]{subfig}
\else
  \usepackage[caption=false,font=footnotesize]{subfig}
\fi
\usepackage{algorithm}
\usepackage{algorithmic}

\begin{document}
%
\title{Affordance Learning In Direct Perception for Autonomous Driving}
%
%
%

\author{Chen~Sun,~\IEEEmembership{Member,~IEEE,}
        Jean~M.~Uwabeza~Vianney,~\IEEEmembership{Member,~IEEE,}
       and~Dongpu~Cao,~\IEEEmembership{Member,~IEEE}
\thanks{C. Sun, J.M. Uwabeza Vianney and D. Cao are with CogDrive Lab in the Department
of Mechanical and Mechatronics Engineering, University of Waterloo, 200 University Ave West, Waterloo ON, N2L3G1 Canada (e-mails: chen.sun@uwaterloo.ca, jmuviann@uwaterloo.ca, dongpu.cao@uwaterloo.ca).}
}

\maketitle

\begin{abstract}
Recent development in autonomous driving involves high-level computer vision and detailed road scene understanding. 
Today, most autonomous vehicles are using mediated perception approach for path planning and control, which highly rely on high-definition 3D maps and real time sensors.
Recent research efforts aim to substitute the massive HD maps with coarse road attributes.
In this paper, we follow the direct perception based method to train a deep neural network for affordance learning  in autonomous driving.  
Our goal in this work is to develop the affordance learning  model based on freely available Google Street View panoramas and Open Street Map road vector attributes.
Driving scene understanding can be achieved by learning affordances from the images captured by car-mounted cameras.
Such scene understanding by learning affordances may be useful for corroborating base maps such as HD maps so that  the required  data storage space is minimized and available for processing in real time.
We compare capability in road attribute identification between human volunteers and our model by experimental evaluation. Our results indicate that this method could act as a cheaper way for training data collection in autonomous driving. The cross validation results also indicate the effectiveness of our model.
\end{abstract}


%
\IEEEpeerreviewmaketitle

\section{Introduction}
 
\IEEEPARstart{I}{N} recent years, autonomous driving technology has become closer to fully being realized. There are many driving forces to this realization, key among them is the advance in perception techniques such as CNN(Convolution Neural Networks). 
The perception techniques allow a self-driving vehicle to understand the driving environment, which is one of the most important steps for vehicle  path planning and control. 
In many applications \cite{janai2017computer}, the autonomous vehicle must be equipped with ubiquitous and robust state-of-the-art vision-based system for it to be able to sense and understand a great deal of driving scene. \par

Recent review \cite{schwarting2018planning} in autonomous driving concluded three main regimes for vehicle planning and control. 
The traditional mediated perception approach sets up the understanding of the road scene by fusing the pre-stored 3-D high definition map and high quality sensor input including LiDAR and GNSS signal \cite{geiger2013vision}. 
The other approach tries to mimic and learn from human expert driving using imitation learning, also called behavior reflex \cite{gaussier1998perception}.  
It's relatively cheap and easier to collect human driving demonstrations at scale. Agents can use the collected data to train a model that maps the sensory and video directly to control outputs without reconstructing the environment.
Dates back to late 1980s, ALVINN\cite{pomerleau1989alvinn} used neural network to construct a direct mapping between camera images and laser measurements to steering output.  Modern works \cite{bojarski2016end, rausch2017learning} deploy DNN control policies using batch imitation learning approach showed the effectiveness in executing simple driving tasks such as lane following. 
Recent research proposed \cite{chen2015deepdriving} the direct perception approach trying to combine the advantages in mediated perception and behavior reflex approaches. The direct perception approach uses neural networks to extract coarse affordances from input video or images and then define mapping from affordances to explicit control. 
As suggested in \cite{chen2015deepdriving}, the mediated perception may add unnecessary complexity to perception problem by detecting redundant objects that may not be useful in driving control decisions. On the other hand, behavior reflex may not be robust enough to adapt to all traffic and driving scenarios. This is due to varying environments and human driving skills. 
In this regard, we agree that direct perception is more robust and efficient and acts as a middle ground between mediated perception and behavior reflex.

	\begin{figure}
		\centering
		\includegraphics[width=0.48\textwidth]{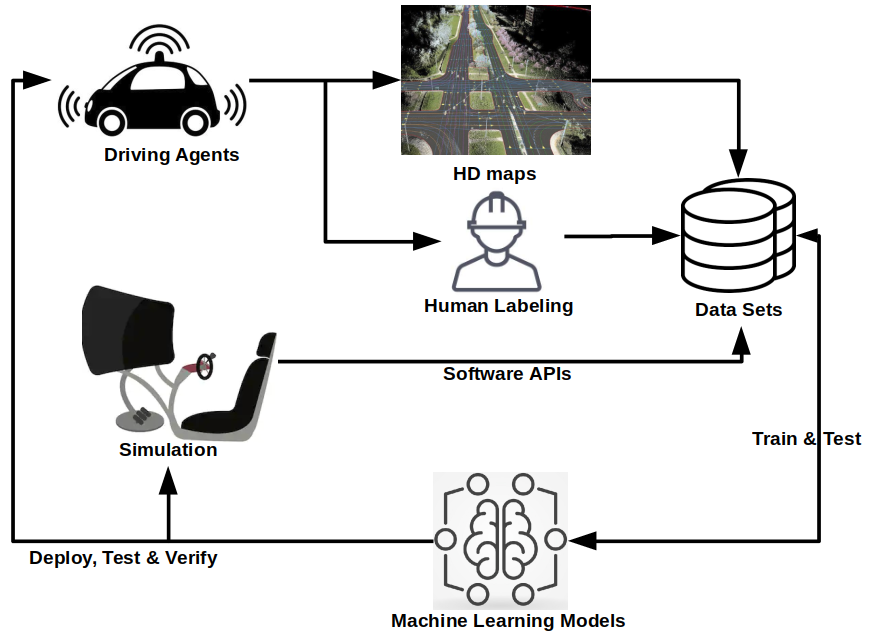}
		\caption{A general pipeline framework for data collection, processing, model training and testing for autonomous driving.}
		\label{framework}
	\end{figure}


There are several challenges in the direct perception approach. 
Since the low level control is decided based on a given set of road attributes, the affordances to be learned must be pre-defined by human. 
Selecting the suitable affordances usually requires feature engineering and driving scene based analysis \cite{silver2010learning}.
After deciding on the coarse road inference layout, we need to collect and label road data in order to develop a relatively good and robust model. 
Ari Seff et al. proposed a method that leverage the google street view images and OpenStreetMap (OSM) for automatic-labelling and model training \cite{seff2016learning}. 
In this paper, we follow the same line with \cite{seff2016learning} for automatic labeling procedure to collect training and testing data. 
Furthermore, we trained a Convolution Neural Network (CNN) to detect traffic scene affordances from a single street view image. 
We have tested effectiveness of the method and accuracy of our model in experiments.
The key contributions of this paper are:
(1) \textit{Efficient CNN Training Model}:
Instead of using pre-trained AlexNet \cite{krizhevsky2012imagenet} CNN model on Places database \cite{zhou2016places} to get good weight as \cite{seff2016learning}, we created a customized CNN architecture based on VGG11 and AlexNet. Using non-initialized model and training on third the number of training images used, we were able to obtain results comparable to \cite{seff2016learning}. (2) \textit{Validation on Automatic labelling}:
We collect data near Waterloo area in Canada while Ari Seff's data are collected in San Francisco, Bay area. We verified their automatic labelling methodology on the different geographical location. Further, in addition to testing our model on the San Francisco GSV images, we examined our network on KITTI \cite{geiger2013vision} tracking data set which is collected in Europe to demonstrate the generalization ability for our model.
(3) \textit{Refining the affordances by driving scene}:
We refine the definition of selected road attributes from \cite{seff2016learning}. 
The affordances set may change according to different driving scenarios.

The rest of the paper is arranged as follows: 
Section II discusses related work in driving context understanding, affordances learning and techniques leveraging open sourced data.
We outline the data collection, affordance selection and auto-labelling methodology in Section III.
In Section IV, we demonstrate the network design and training methodology compared with recent research works. 
Experimental results and discussion are shown in Section V, along with the conclusion and future work discussion in Section VI.

\section{Related Work}
 
Traffic scene and driving context understanding are an ongoing challenges in autonomous driving. 
Over the past few years, more of more resources and focus have been put towards scene understanding as a primary challenge in autonomous driving, especially since DARPA Urban Challenge \cite{buehler2009darpa}. 
One type of strategy for the static driving context understanding is simultaneous localization and mapping
(SLAM) \cite{choudhary2014slam}. 
A virtual representation of the road, traffic and surrounding buildings can be constructed based on the 
reconciliation of real-time sensor data and pre-stored HD maps.  
With the detailed driving context representation, the detailed path planning and driving policy can be further derived. 
However, the main bottlenecks for this type of approaches are the high requirements on computing power
and data transmission \cite{seif2016autonomous}.

Vision based methods tries to mimic the human driver using camera recorded images as major sensory input. German Ros et. al presented an Offline-Online perception framework in \cite{ros2015vision} where the 3D semantic maps are pre-stored offline and online semantic segmentation can be achieved by performing SVM based classification on video-sequences. 
While re-localization process in this framework can be achieved real-time, the online retrieval of semantics does not necessarily adapt to environment change.
Authors in \cite{teichmann2018multinet} proposed a unified multi-net structure that performs the joint classification, detection and semantic segmentation in real time. 
Such driving context understanding methods like  semantic segmentation with camera images are eventually aim to assist the control design for the ego vehicle. 
The research group in Princeton University demonstrated the idea of directly learning the affordances from an image using the direct perception approach \cite{chen2015deepdriving}. 
They  train images (from a car racing game TORCS) using a ConvNet to predict affordances such as host vehicle distance to front vehicle or left/right lanes for driving action.
They tested their approach both on virtual and real environments and reported good performance in close range to the state-of-the-art deformable parts model car detector \cite{geiger20143d}. 
Based on the determined affordances, they are able to build simple rule based controller for the vehicle control in TORCS. This idea proves that meaningful driving affordances can be incorporated in autonomous driving decision making.
Similar approach was adopted by \cite{sauer2018conditional} where they further advanced the driving image affordance learning and control with conditional neural network and more photo-realistic simulation platform CARLA \cite{Dosovitskiy17}.

Recent research \cite{caltagirone2017simultaneous, xing2018advances, xu2017end} demonstrated that data-driven perception model often surpass the hard-coded reasoning in context prediction leveraging the large-scale data since much more expert driving experience can be exploited. 
However, the process of data-set collecting and labelling often requires huge amount of effort. 
Typical pipeline framework for data-driven approach can be concluded as Fig. \ref{framework}.
The aforementioned research works \cite{chen2015deepdriving, Dosovitskiy17, sauer2018conditional}  use simulation data since the ground truth information is programmed and can be exported through provided APIs. 
However, there is still a gap between the simulated environment and the real world data \cite{wang2017parallel}. 
Many open-sourced driving data sets received increasing attention in the recent years.
The Caltech lane data set \cite{aly2008real} focused on lane marking wheares KITTI \cite{geiger2013vision} provide detailed marking with multiple sensors including LiDar. 
With high-end expensive sensors the vehicle dynamic state estimation could be achieved by methods mentioned in recent review paper \cite{guo2018vehicle}.
Recenly, Xu et. al published BDD100K data set \cite{xu2017end} where the diverse driving data is collected in a distributed  way by Uber drivers across California and New York and annotated by human labour.
These data sets were collected and labelled with a deliberate designed system but are not automated and still quite expensive. 
OpenStreetMap (OSM) \cite{haklay2008openstreetmap} is an open-source mapping project started since 2004, where over 21 million miles of road geographical information is available for public use. 
In \cite{seff2016learning}, the authors trained a CNN model to predict road attributes using images from Google Street View (GSV). They presented automatic labeling method based on location matching with attributes from OSM. 
We follow the similar idea with \cite{seff2016learning} where we use 'cheap' data with automatic labelling to teach a model predicting cues given image inputs. 
In our work, we refine road attributes and how it influences driving actions, using not only intersection but also affordances on a straight road.

\section{Dataset and Labelling}
As we have discussed in previous section, deep ConvNet frameworks was used to train data and determine affordances such as host vehicle to road angle, number of lanes and driveable areas in many recent research works. 
However, this requires huge number of labeled images to be able to train a reliable model. There are a number of real world street scene labeled data sources such as KITTI \cite{geiger2013vision} and synthetic data (from games and movies) such as Virtual KITTI \cite{gaidon2016virtual}.
For tasks such as determining bike lanes, wrong-way vs. right-way, the available data in \cite{geiger2013vision, gaidon2016virtual}  and most other open source autonomous driving benchmark data sets \cite{xu2017end} may not be enough or labeled for these specific affordances.
Consequently, as proposed in  \cite{seff2016learning}, we take advantage of huge free and open source imagery and corresponding attributes repository available on GSV and OSM, to train a ConvNet model to predict the road attributes.

\begin{figure}[h]
\centering
\subfloat[]{
\includegraphics[width=0.19\textwidth]{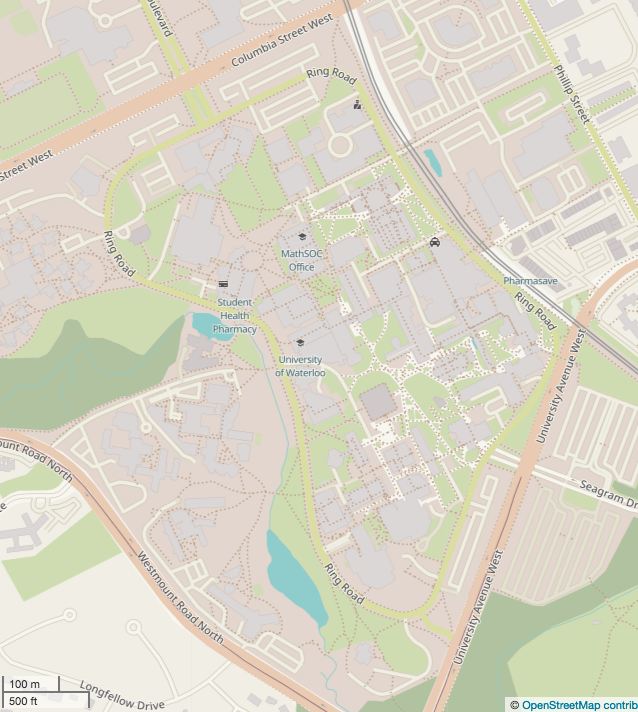}
\label{fig:subfig1}}
～
\subfloat[]{
\includegraphics[width=0.26\textwidth]{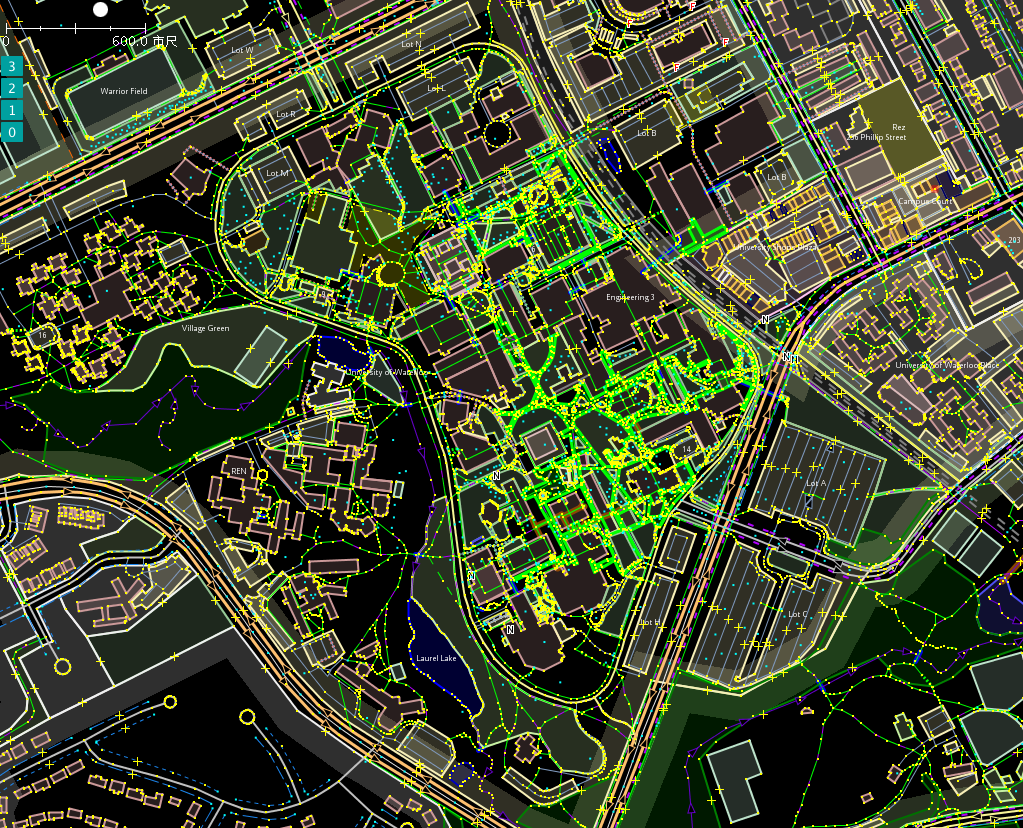}
\label{fig:subfig2}}
\caption{A standard OSM map covering area near University of Waterloo in Ontario Canada is shown in (a). The same area highlighting high density of attributes available in OSM such as road polylines, parcels near the road and building polygons are shown in (b). Each node in the map contains coordinate information that can be used to associate it with other location based features from sources such Google Street View panoramas. }
\label{gsv and osm}
\end{figure}
\subsection{Data Collection}

OpenStreetMap is a community driven and local knowledge based open data platform. The contributed data is tied together using location information in a world geodetic coordinate System (WGS84). Similarly, Google has a huge deposit of street view imagery with each panorama encoded with vehicle true heading at the time of image capture and location information in WGS84 coordinate system. Using location neighborhood constraint, it becomes possible to associate each panorama from GSV with nearby road attributes such as number of lanes or if an intersection is likely in view \cite{haklay2008openstreetmap}. The accuracy of feature association is directly affected by the location accuracy in both GSV and OSM and whether information in both sources were updated in same time frame. As will be highlighted later, we found some mislabeled affordances due to time latency and unresolved location differences especially at bridges or close road networks where a small location deviation would associate features of one road to an image showing a different road.   

The following sections give more details about the process.

\subsubsection{Google Street View Panoramas}

\begin{figure}
		\centering
		\includegraphics[width=0.46\textwidth]{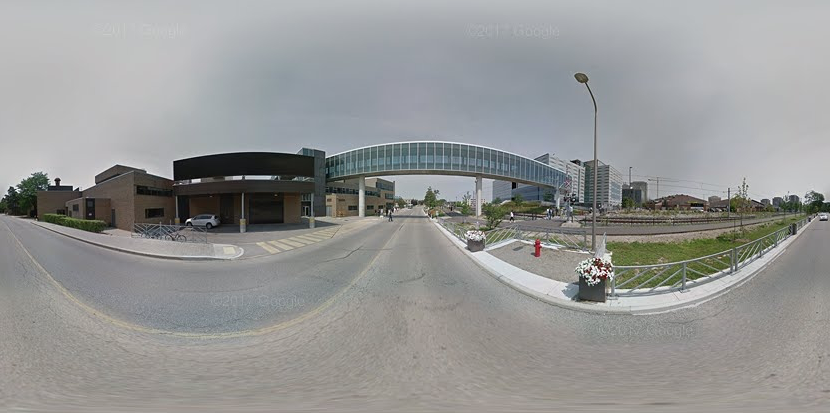}
		\caption{A seed panorama used as a starting point for downloading GSV panoramas. The pedestrian bridge in view connects Engineering buildings 3 and 5 (E3 to E5) at University of Waterloo.}
		\label{uwaterloo}
	\end{figure}

We have downloaded over one hundred thousands panorama images starting with a seed panorama image ID at University of Waterloo (shown in Fig. \ref{uwaterloo}). These panoramas were then cropped and warped into $227 \times 227 \times 3$ sized images with a field of view (FOV) of 100 degrees. 
The image size and FOV were kept similar to what \cite{seff2016learning} used after finding them sufficient for driving scene. Each image is encoded with coordinates in WGS84 reference system. This is later used to query and overlay with data from OSM.

\subsubsection{OpenStreetMaps Vectorized Data}

OSM \cite{haklay2008openstreetmap} is a vectorized map with attributes contributed by volunteers. Attributes include but not limited to poly-lines such as those defining extents of road networks, bike lanes and traffic markings. It also includes point features such as stop signs, traffic lights and speed signs. Its data availability may be lacking in rural areas or small towns, since volunteers tend to contribute to maps around where they reside.

\subsection{Affordance Labeling}

Here we discuss the affordance set selection. It is still an open question nowadays for the optimized  road attributes  selection for the driving context understanding. 
Chen et al. proposed 13 affordance indicators in \cite{chen2015deepdriving} for multi-lane tracks in TORCS. 
It is rather simplified since there is neither intersection, pedestrian nor traffic light in TORCS. 
Authors of \cite{sauer2018conditional} advanced the affordance learning in the single lane urban scenario simulated in CARLA where 6 affordances were selected. 
Both works utilized the global information embedded in the simulation engine such as the global information for all the agents in the map and the distance for the vehicle between the road center line. 
It is rather difficult for us to obtain these global information in real data, hence we choose the target road attributes based on the available OSM and GSV data. 
The OSM  data set and GSV panoramas are encoded with location coordinates in WGS84 reference frame. It was possible to query and overlap an image cropped from GSV panoramas with corresponding attributes in the OSM data. 
We conclude the list of automatically labeled affordances in TABLE. \ref{attributes}.

\begin{table}[htbp]
\centering
\caption{Road Attributes labeled for waterloo region GSV images} \label{attributes}
\begin{tabular}{|l|l|l|}
\hline
\textbf{Labeled Affordances} & \textbf{Data Type} & \textbf{Range}  \\ \hline
Heading\_Angle               & Continuous         & $[-\pi, \pi]$   \\ \hline
Drivable\_Heading?           & Boolean            & \{True, False\} \\ \hline
Intersection\_Ahead?         & Boolean            & \{True, False\} \\ \hline
Distance\_to\_Intersection   & Continuous         & {[}0, 30{]} m   \\ \hline
Number\_of\_lanes            & Discrete           & \{1, 2, 3\}     \\ \hline
Wrong\_Way?                  & Boolean            & \{True, False\} \\ \hline
Bike\_Lane?                  & Boolean            & \{True, False\} \\ \hline
\end{tabular}
\end{table}

\subsubsection{Lane Following}
\begin{figure}[htp]
		\centering
		\includegraphics[width=0.48\textwidth]{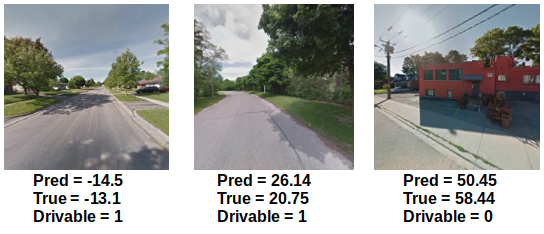}
		\caption{Vehicle heading angle ($^\circ$) prediction result and drivable classification. }
		\label{heading_drivable}
	\end{figure}

The vehicle heading angle is rather important for lane following task. We label the affordances \textbf{Heading\_Angle} to represent the current ego vehicle heading angle corresponding to the driving lane. Fig. \ref{heading_drivable} gives examples of the angle prediction using our model. 
We further extend the angle prediction to a classification problem in order to compare human capability of identify the heading angle from single image.
The detailed comparison will be discussed in Section V. 

The \textbf{Drivable\_Heading} is to detect whether there is a continual path to execute the lane following  task. 
In \cite{seff2016learning}, this attribute is used to predict whether the current heading of the panorama image align with the road disregarding the driving lanes. In our work, we refine the definition of \textbf{Drivable\_Heading} where not only the \textbf{Heading\_Angle} is within $22.5  ^{\circ}$, there should not be any static object in front of the panorama (road edges, endings). 
Some examples of our model prediction and calculated true labelling are demonstrated in Fig. \ref{heading_drivable}.
The labelled vehicle heading angles are calculated based on OSM lane attributes and GSV panorama applied rotations. Note that the true labels are not necessarily accurate since the calculation is based on one assumption that the vehicle that collected the GSV panoramas are driving heading straight in most of the lanes.
Nevertheless, we found that our model predictions on heading angle trained with these generated labels showed good level of effectiveness. We will discuss more on this issue later in Section V.
 
\subsubsection{Intersection Handling} 
\begin{figure}[htp]
		\centering
		\includegraphics[width=0.48\textwidth]{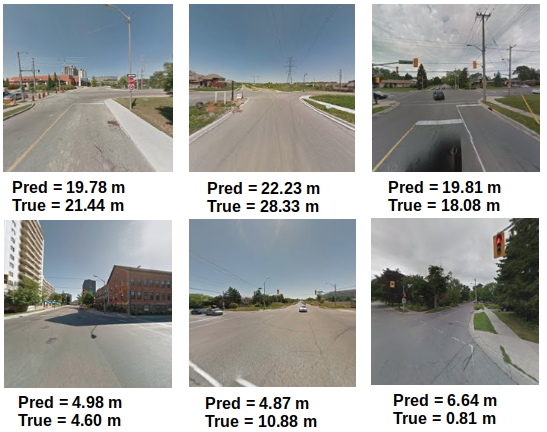}
		\caption{Comparison between prediction and true labels on distance to intersection. The `true' distance label is calculated by measuring the distance between the GSV referring point  and the center of the intersection in OSM.}
		\label{intersection}
	\end{figure}

Various intersections is the most common scenario in the urban driving setting where human driver need to decide the high level command in order to follow the planned driving path. 
We denote the the following two road attributes for the intersection handling: \textbf{Intersection\_Ahead?} and \textbf{Distance\_to\_Intersection}.
We follow the similar procedure with \cite{seff2016learning} to mapping the intersections where first locate them on OSM by finding shared vertices and then mapping with GSV panoramas. The nearest intersection distance from the corresponding ego vehicle heading is used as our ground truth labelling. 
We use the GSV images and the computed ground truth images to train our model for the distance prediction as we can see in Fig. \ref{intersection}.
The \textbf{Intersection\_Ahead?} is a binary classification problem. After the manually comparing and assessing with the sampled images, we agreed with the methodology in \cite{seff2016learning} where we choose given distance 30m  and below as the exist intersection ahead. 
Similarly, we choose those intersection distance greater than 100m as false prediction in order to provide clear distinction for the classification task.

Two set of sample results are demonstrated in Fig. \ref{intersection}. It is an obvious task to identify or measure the distance between intersection when you approaching one, as we can see from the top three samples in Fig. \ref{intersection}.
However, the estimation error of our model grows when predicting on a view  at the intersection. 
The visual inputs at intersections usually are not as structured as general road segments. The open view of a unstructured terrain usually confuse CNN based model since only one shot of the image is given. 
We believe the prediction results can be improved by using memory based model such as LSTM \cite{gers1999learning}. 

\subsubsection{Multi-lane Handling} 
\begin{figure}[htp]
		\centering
		\includegraphics[width=0.48\textwidth]{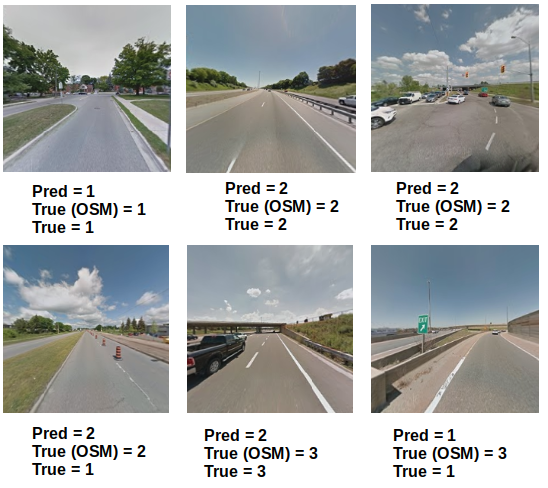}
		\caption{Multi-lane prediction using our model trained on labels provided by OSM. In some cases when there is a dynamic change (construction, road change etc), static labels from OSM 
		cannot represent well of the road segment (Bottom left \& right).
		The narrow view and obstruct by dynamic objects (Bottom middle) may also result in false prediction. 
		}
		\label{num_lanes}
	\end{figure}

It is rather important for autonomous driving can identify the multi-lane driving context especially in urban or highway driving in order to proceed to path planning and driving maneuver control.
The attribute \textbf{Number\_of\_Lanes} identify how many lanes in the current driving road.
It is addressed in \cite{seff2016learning} where they only include one-way roads in training data due to the inconsistency for two-way roads when considering the driving direction. 
In our work, we further included the images of two-way roads in our training set.
We find that our model prediction was consistent with the labels for the most part, except when there was occlusion, lanes were not visible or the label was incorrect.
In Fig. \ref{num_lanes}, top three images demonstrated the effectiveness of our model prediction. Despite the curved road shown in top right image, the model was able to generalize well and made correct prediction.
Yet the task for predicting number of lanes from single shot of image input is still challenging as pointed out by  \cite{seff2016learning} due to the lack of lane markings.
We list three typical false prediction in the bottom of Fig. \ref{num_lanes} as a complement of \cite{seff2016learning}.
The road constructions or other dynamic changes may result in consistency between expected truth of the driving context and the static road attributes labels. The static OSM data cannot adapt to recognize the traffic cones as demonstrated in the bottom left of Fig. \ref{num_lanes}. 
Furthermore, the GPS location accuracy may raise the label error issue especially near intersections or highway ramps (bottom right in Fig. \ref{num_lanes}).
In general, our model can predict well on the number of lanes given enough lane markings or other vehicles. Unfortunately, the dynamic change of the road segments and obstruction of the view may result in false predictions.

\begin{figure}[htp]
		\centering
		\includegraphics[width=0.48\textwidth]{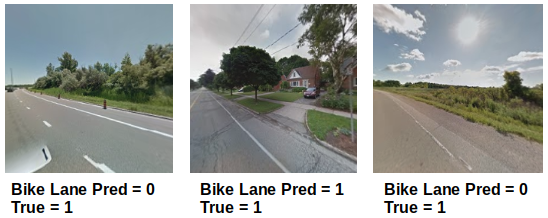}
		\caption{Exist bike lane prediction using our trained CNN model.}
		\label{bike_lane}
	\end{figure}

One side problem for counting the number of lanes is to identify the possible road side bike lanes. The road attribute \textbf{Bike\_Lane?} is true if there exist bike lane based on the given panorama. 
Our trained model  to predict bike lanes performed 3\% worse than Seff et. Al's in \cite{seff2016learning}. However, It should be pointed out that the validation accuracy was affected by the mislabeled images. As explained in the previous results, in some cases, the models made correct predictions despite of incorrect labels. For bike lanes, this is still the case. As can be seen in Fig. \ref{bike_lane} on left image, the model predicted the road to have no bike lanes and this is correct from visual inspection. However, the image label indicated that there is a bike lane. It is likely that the previous views of the road had bike lanes but ended before intersection. After examine on the false prediction samples we find that the sampled bike lanes tend to end before major intersections. 
Bike lane may confused with the highway emergency lane, due to the CNN model only take single shot image as input. One false prediction example is given in Fig. \ref{bike_lane}  on the right where the model may treat rural road with a bike lane as the highway ramp or emergency lane.

\begin{figure}[htp]
		\centering
		\includegraphics[width=0.48\textwidth]{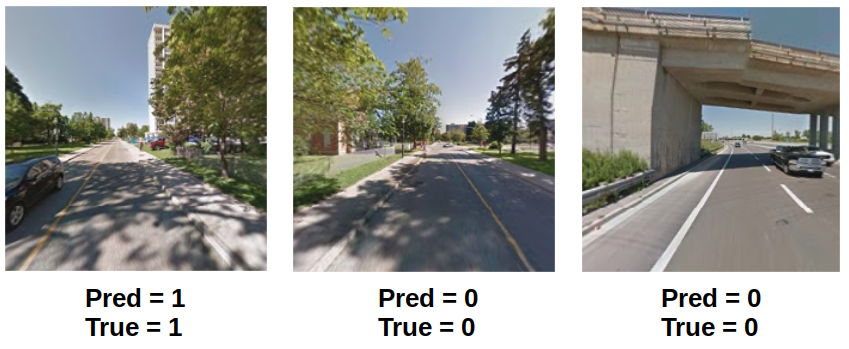}
		\caption{Right or wrong way classification using our trained CNN model. The right of way corresponding to label 1 and wrong way is labelled as 0.}
		\label{rwways}
	\end{figure}

We also interested to identify whether the ego-vehicle is on the right way of driving given the panorama image.
\textbf{Wrong\_Way?} classification is based on the driving rules that you must drive on the right side of the road. 
By carefully examining left and middle images in Fig. \ref{rwways}, one can verify that indeed the model makes correct predictions. For image on the left, it is correctly predicted to be the right-way. This is informed by the driver view largely being on the right side of the road. The middle image is classified as wrong way of driving, which is correct since the driver view mostly falls on the left side of the road. Image to the right of Fig. \ref{rwways} is less ambiguous to classify and correctly predicted as wrong way.  

\section{Model Design and Training}

\begin{figure}[htp]
		\centering
		\includegraphics[width=0.48\textwidth]{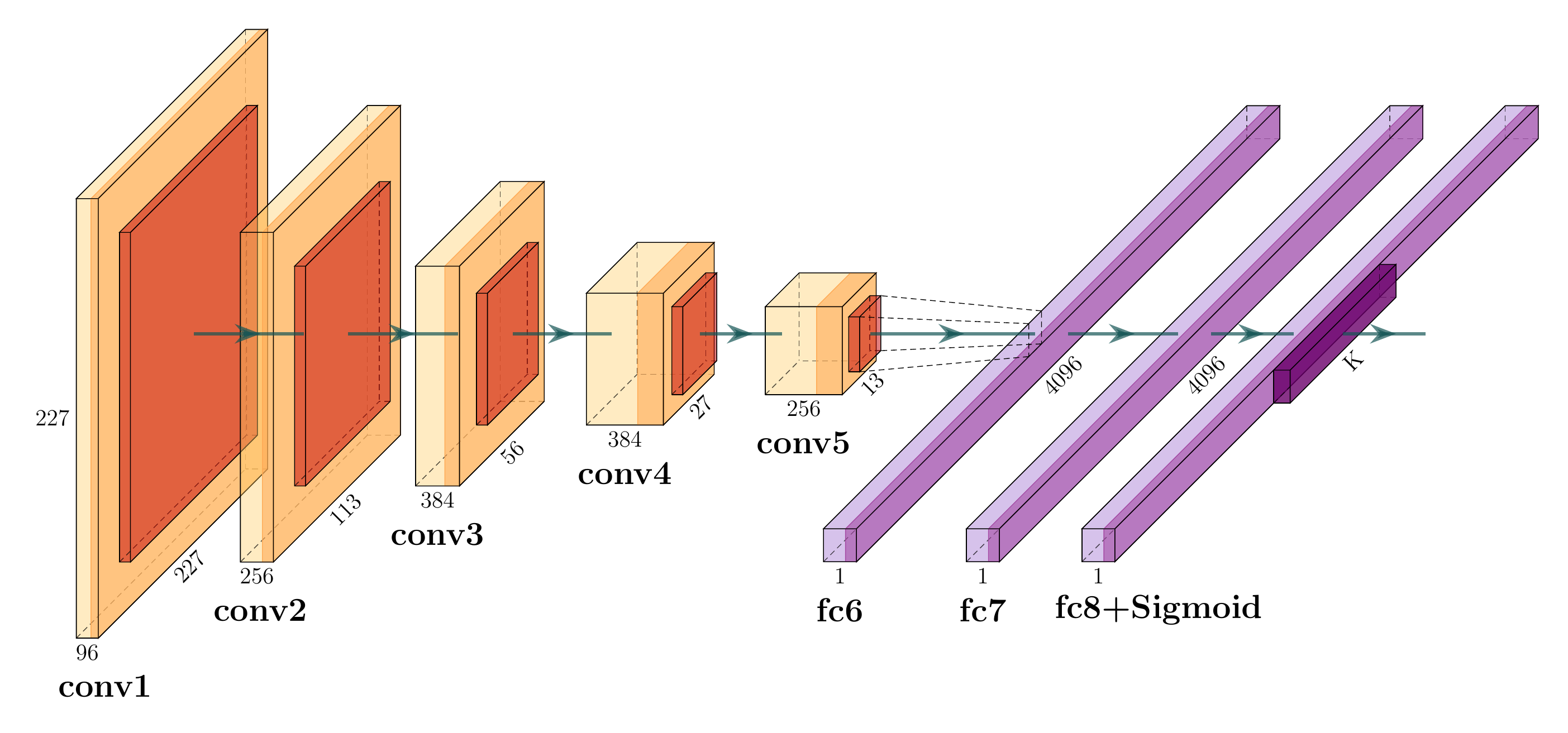}
		\caption{The architecture of the proposed CNN. The inputs is the crop of GSV panorama image and the output layer consists of selected features and affordance indicators. 
		Note that we perform batch normalization  after convolution layers 1, 2 and each fully connected (FC) layers instead of dropouts to reduce over fitting issue.}
		\label{network}
	\end{figure}

Initially, we followed the same path as \cite{seff2016learning} using AlexNet \cite{krizhevsky2012imagenet} architecture for our model. 
However, we tuned hyper-parameters to find the architecture that is most efficient and produced best validation accuracy for driving affordance training. 
As shown in Fig. \ref{network}, we ended up with a network comprised of five Convolution layers as AlexNet and three fully connected (FC) layers each with 4096 channels. 
We found that using $3 \times 3$ receptive field for each convolution layer as proposed \cite{simonyan2014very}, produced better validation accuracy for all trained affordances. 
For all convolution and FC layers we used rectified linear (ReLu) as activation function. 
We also applied padding and max pooling to preserve input size and spatial resolution, respectively,  through convolution layers. We chose not to use dropouts but instead employed batch normalization as a substitute, after the first two convolution layers and  each FC layer. This increased robustness of the weights and reduced over fitting while also preserving the learned features. Higher weights at each layer (except output layer) were penalized by applying an L2 regularization of 0.0001. This helped us keep an eye on learned weights and drastically reduced over-fitting. The output layer structure depends on whether the model is for regression or binary classification. For regression, we used an output layer with one kernel and no activation function, i.e. the outputs were not scaled into probability output. The model was compiled using RMSprop optimizer with a learning rate of 0.0001 and mean squared error (MSE) as the loss function. The accuracy of our regression model was reported in mean absolute error (MAE). The output layer for binary classification model has one kernel and uses sigmoid as the activation function to scale the predictions into values between 0 and 1. Similar to regression, the compiling was done using RMSprop optimizer with learning rate of 0.0001. However, the loss function was changed to binary crossentropy with accuracy reported as a percentage.  

Our model was built in Keras \cite{chollet2015keras} running on top of TensorFlow framework.  As shown in Fig. \ref{network}, the first convolution layer input accepts an RGB image of size $227 \times 227 \times 3$ and passes it through 96 filters of size $3 \times 3$ with ReLu as activation function, strides of 1 pixel and padding set to 'same' i.e., it outputs same dimensions as input. Kernel regularizer is set to L-2 norm. This is followed by a scaled and centered batch normalization layer with 0.99 momentum and 0.001 epsilon. The moving variance initializer is set to one. A $3 \times 3$ max pooling layer with strides of two pixels and padding set to 'valid' (no padding), comes next. 
The second convolution layer has 256 filters of size $3 \times 3$. The activation function, stride, padding and regularizer are set similar to first convolution layer. The batch normalization and max pooling layers follow (with similar set up as previous described). The 3rd and 4th convolution layers have 384 filters of size $3 \times 3$ and are separated by a max pooling layer. Another max pooling layer is inserted before the 5th convolution layer with 256 filters of size 3x3 (all convolution layers maintain similar structure to first layer. they only differ in number of filters). A flattening layer is implemented before the first FC layer. All the fully connected layers have 4096 neurons with ReLu as the activation function and L-2 norm (0.0001) regularizer. Each of them are separated by a batch normalization layer.

Before training the images were normalized by changing the pixel values to float and diving with 255. random images were augmented by applying a rotation of $22 ^\circ$, width and height shift of 0.2, shear of 0.2 and zoomed by a factor of 0.2. The images were trained in batches with a batch size of 32 and 50 epochs. The steps per epoch for both training and validation depended on total number of images as   in Eq. \eqref{stepsperepoch}. Images and corresponding labels were also randomly shuffled during the training phase.
\begin{align}\label{stepsperepoch}
 StepsPerEpoch = \frac{Number~of~Images}{Batch~Size} 
\end{align}

As a comparison with \cite{seff2016learning}, their model was pretrained on Places Database
while ours used random initialization without pre-trained weights. Their model tend to converge after a ``few thousand iterations". Instead, our model used only 50 epochs and less than two hundred steps per epoch since we use much smaller data set.  The detailed model performance comparison and cross validation will be given in next section.

\section{Results and Discussion}

In this section, we present the quantitative evaluation result of the proposed model.
We also  discuss the improvement on our model and current data collection pipeline for autonomous driving.

 \subsection{Accuracy Evaluation}
  We validate our CNN models  performance across three different geographical regions, namely data collected from Waterloo (abbreviated as W), data used in \cite{seff2016learning} collected in San Francisco, Bay area (abbreviated as SF) and KITTI tracking data collected from Europe. We also provide the comparison between human baseline and model prediction on classification tasks.

  \subsubsection{Our CNN models vs. Human}
\begin{figure}[htp]
		\centering
		\includegraphics[width=0.48\textwidth]{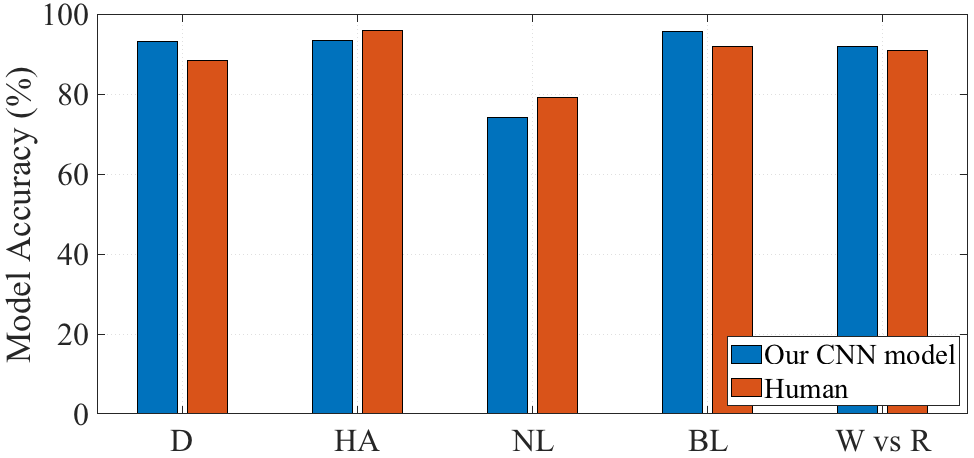}
		\caption{Comparison between human baseline and our trained CNN model on the classification prediction accuracy (higher is better). The tasks investigated here are drivable (D), heading angle (HA), number of lanes (NL), bicycle lanes (BL) and right way or wrong way prediction(W vs R).  }
		\label{human}
	\end{figure}

We asked five human volunteers to label 1000 images for each affordance. We evaluate our models on the same images and compare results which are presented in Fig. \ref{human}.
We focused on classification tasks as we found it difficult for humans to meaningfully measure angles or distances from a low quality images. 
Consequently, we did not consider distance to an intersection and heading angles were deduced to binary classification by asking humans to predict whether the image showed a negative rotation (left rotation with respect to road) or positive rotation (right rotation with respect to road).

Each human volunteer was first shown ten example images and corresponding labels for each affordance under consideration, in order to train them and highlight the image to affordance association in the context of driving. We then let each volunteer label provided images per affordance without access to OSM derived true labels. 
Consequently, for each affordance, we generated a single set of human labels by combining five individual labels using consensus model \cite{herrera2002consensus}. 

\par As evident in Fig. \ref{human}, our CNN model predictions and human labels were within $\pm 5.8\%$ of each other. Our model performed better than human for driveable space (D), bikelane (BL) and wrong-way vs. right-way (W vs R) affordances. Also, it had comparable results for number of lane (NL) and driving heading angle (HA) affordances.

  \subsubsection{Model Generalization Test}

\begin{figure}[htp]
		\centering
		\includegraphics[width=0.48\textwidth]{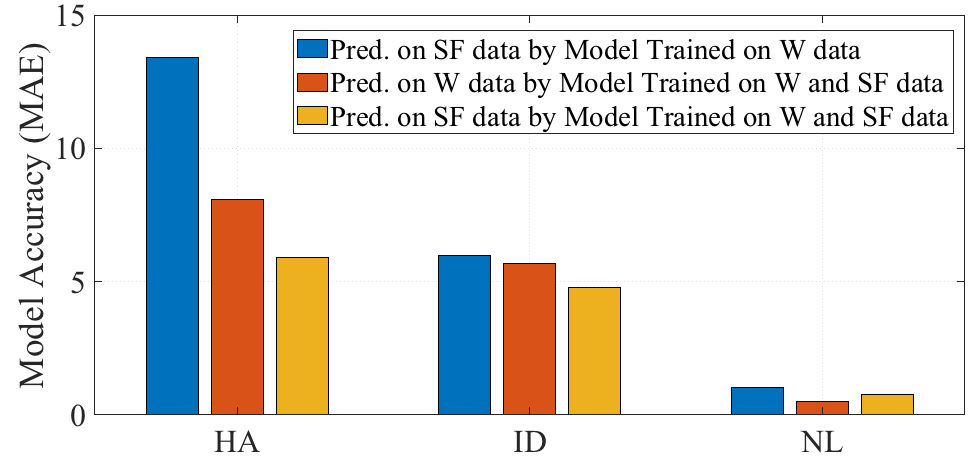}
		\caption{Comparison of mean absolute error between our CNN model trained and tested on data sets across different geological regions (lower is better). The tasks investigated here are heading angle (HA), intersection distance (ID) and number of lanes (NL). }
		\label{geological}
	\end{figure}

  To find out how well our model would generalize on data collected in different geographical locations, we took advantage of GSV panoramas from San Francisco Bay Area available for download on \cite{seff2016learning} data page. 
  We did cross validation by comparing the prediction on San Francisco Bay area GSV images by model trained on Waterloo dataset, prediction on Waterloo GSV images by model trained on both Waterloo and San Francisco datasets and prediction on San Francisco Bay area GSV images by Model trained on Waterloo and San Francisco data sets.
  We then used the CNN models for heading angle (HA), intersection distance (ID) and number of lanes (NL) affordances that were trained on Waterloo data set (from henceforth referred to as model set 1 to predict on San Francisco Bay Area images. We recognized that data augmentation applied to training data might not be robust enough for driving scene training since it is relatively hard to augment buildings and other features (such as trees, grass and curbs) proximity to road. Hence, we trained new models for HA, ID, and NL using a data set with half of the images from San Francisco and another half from Waterloo (referred  as model set 2. This model was tested on a Waterloo data set and San Francisco data set, independently. We were careful to make sure that the test images were never used during training and validation of the models. However, the same testing data set from SF was kept consistently the same in model set 1 and model set 2 testing. The MAE between the predictions and true labels were computed and plotted in Fig. \ref{geological}. The MAE plot shows that model set 2 is more accurate and generalizes better than model 1. 
  Model set 2  performed best in all affordances. We should also point out that the difference in MAE for both models should be examined independently for each affordance. 
  For instance, number of lanes in Waterloo range from 1 to 4 lanes. Therefore a MAE of 1.04 in number of lane would be considered too big since it means that a model would likely be predicting wrong number of lanes most of the time. However a MAE of 4.8 meters for intersection distance may be tolerable  given that the intersection distance in consideration range from 0 to 30 metres.
  
  \par In Table. \ref{table-comp}, we compare the the performance of the proposed architecture and trained models by making prediction on San Francisco testing data. Three set of model are compared. We use the model proposed in \cite{seff2016learning} trained in SF data as an baseline. The other two model are aforementioned model 1 and 2 using the same architecture. 
  We compare the accuracy relative to the training data size used in HA, ID and NL affordance training. 
  Our models trained on both Waterloo and San Francisco datasets performed better than models in \cite{seff2016learning} for HA and NL affordances. This is despite only using $1/3$ of the data size. 
  We used 4K images while \cite{seff2016learning} used over 12K images for training.  Although our second model trained on just Waterloo (W) datasets while it reports worst accuracy, it is still comparable to Seff et Al. results for all the three affordances.
  Moreover, the model trained using only 4K images on the combined data outperforms the other models in most of the regression tasks. This agrees with most of the findings in data augmentation where it is best to use data collected in various geographical locations and in different conditions in order to train a perception model that could generalize well.
  
  	\begin{table}[htp]
\centering
\begin{tabular}{|l|l|l|l|}
\hline
 &  Model in \cite{seff2016learning}& Ours on W & Ours on W \& SF \\ \hline
Train Samples & \textgreater 12K & $\sim$6K & $\sim$4K  \\ \hline
ID  (MAE) & 4.3 & 6.01 & 4.77 \\ \hline
HA (MAE) & 9.2 & 13.4 & 5.89 \\ \hline
NL (MAE) & 0.9 & 1.04 & 0.76 \\ \hline
\end{tabular}
\caption{The comparison across models proposed in \cite{seff2016learning} trained based on SF data, our architecture trained on W data as well as the same architecture trained on combined SF and W data. All the models compared here are tested on same SF data. Our model trained on W and SF data reports best results. }
\label{table-comp}
\end{table}

  \subsubsection{Driving Heading Angle prediction on KITTI dataset}
  
\begin{figure}[htp]
		\centering
		\includegraphics[width=0.48\textwidth]{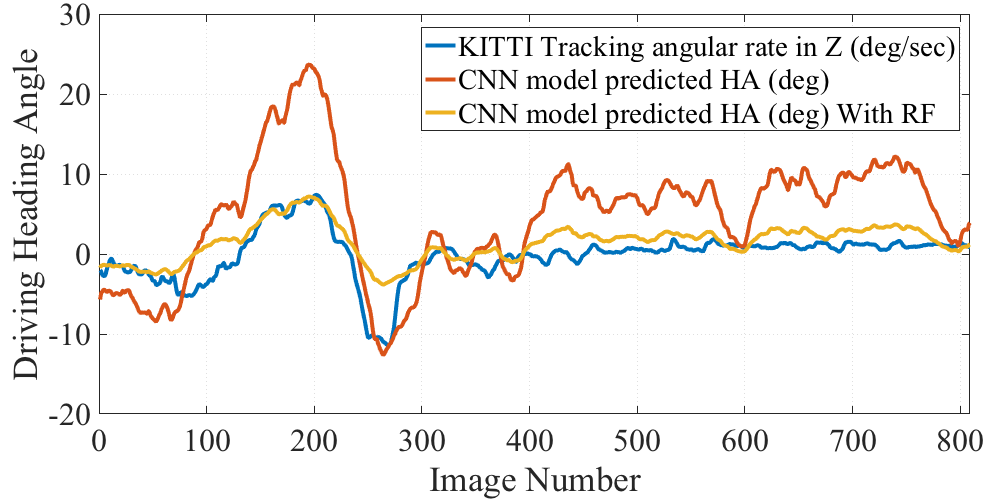}
		\caption{Comparison on regression task of vehicle heading angle prediction.
		We feed the resized KITTI images (collected in Europe) into our CNN model trained  on data collected in Waterloo area in Canada. The blue line corresponds to the ground truth measurement from KITTI, and the red line corresponds to the raw prediction result from our CNN model without considering the distortion of the input image. The CNN prediction with applying the resizing factor is plotted in yellow.
	}
		\label{kitti_compare}
	\end{figure}

   Our driving heading angle model was used to predict on KITTI tracking data set and results are plotted  in Fig. \ref{kitti_compare} with the angular rate around Z axis for each image as included in the image metadata from KITTI website.
   This plot clearly shows similar trend between our CNN model predicted heading angle with reported angular rate at the time of image capture. 
   Note that the size of the  KITTI tracking images are $1242\times375$ in width and height, however our trained model takes $227\times227\times3$ input. 
   It is hard to resizing the KITTI images in scale to fit the input size of our model without cutting out any road features.
   Unfortunately, we lose spatial resolution and apply distortion by shrinking the image horizontally. We can observe the magnitude difference between the red line (raw CNN model prediction) and the blue line (KITTI ground truth).
   In order to demonstrate this issue, we applied the resizing factor (RF) plotted the new heading angle magnitudes. As evident in Fig. \ref{kitti_compare}, the heading angles with resizing factor applied are very close to the KITTI tracking angular rates at the image capture.

  \subsection{Automatic Labeling for driving data sets}
 
  The automatic labelling for driving data by leveraging existing OSM and GSV data is a complement way or cheap substitution of generalizing training data workflow for autonomous driving. The growing use of OSM data for training may contribute to more accurate static labelling in return.
  Furthermore, these road attributes can be stored as a light reference vector map with very fast accessing speed for autonomous driving as a complement of heavy and expensive high-definition maps. 
  
  It is rather important to increase the accuracy of automatic labelling as we have demonstrated in previous sections. 
  The correctness of automatic labeling was defined by a number of factors. First, positioning in both GSV and OSM data carries a degree of error.
  GSV panoramas are collected using Google Street cars equipped with navigation system (GPS/INS) whose accuracy depends on environment. 
  Its positioning accuracy in these areas can range in meters. When GPS and IMU are combined to create a fused solution, a centimeter-level accuracy (after post-processing or using real-time kinematics) can be achieved. 
  However, this is true in open sky areas as GPS signal is easily obstructed in areas with a lot of buildings or trees causing deterioration of accuracy to decimetre-level accuracy even with a high-end IMU \cite{zhou2015new}. These positioning challenges are inherited by the collected panoramas and contribute to mismatching with OSM data. Another factor is that OSM data is contributed by volunteers and hence integrity of its data varies and may not be up-to-date. This is highlighted in Fig. \ref{bike_lane} where the left image is labeled as having a bike lane but in reality it is the road segment prior to current location. Moreover, the OSM road attributes data and GSV images are static which means that they cannot represent well when there is dynamic change of the road segments such as road construction, change of weather etc. 
  The road construction may affect the correctness of the labelling more as shown  in the bottom left image of Fig. \ref{num_lanes}.
  The GSV panoramas are collected by google street car mostly in sunny days with clear view with almost no variation on weather.
  The driving data from different weather such as raining and snowing are necessary in order to train a robust perception model that could generalize well. 
  The prediction error may also inherited from the downside of the CNN architecture where only single shot of front view image is used as input, which is presented in the bottom middle image of Fig. \ref{num_lanes} when there is dynamic obstruct blocking the view. 
  Designing a distributed way of collecting visual driving data across various countries and under various weathers and applying network structure with memory block and using
  images sequence as input for prediction are both potential future improvement for this line of work.

\section{Conclusion}

In this paper we proposed an efficient CNN model for driving affordances learning by leveraging online static databases. The open resourced road data is collected in a distributed way across the world which enable us automate the data collection and labelling pipeline.
We have examine our trained model based on different data set across geographical regions. The quantitative results indicated the effectiveness our CNN model for affordance prediction across driving data collected in Waterloo area in Canada, California area in US and Europe respectively. The aim of this work was to extend the automated pipeline approach for training affordance learning. Exploring more advanced neural network structures and refining the static OSM labels with dynamic observations is currently under investigation.

\ifCLASSOPTIONcaptionsoff
  \newpage
\fi



%
\bibliography{affordance_manuscript}{}
\bibliographystyle{IEEEtran}


%

%








\end{document}